\documentclass{article} 
\usepackage{iclr2015,times}
\usepackage{hyperref}
\usepackage{url}
\usepackage{amsmath,amsthm,amssymb}
\usepackage[pdftex]{graphicx}
\usepackage{epstopdf}

\title{Deep metric learning using Triplet network}

\author{
Elad Hoffer \\
Department of Electrical Engineering\\
Technion Israel Institute of Technology\\
\texttt{ehoffer@tx.technion.ac.il} \\
\And
Nir Ailon \thanks{The author acknowledges the generous support of ISF grant number 1271/13}\\
Department of Computer Science\\
Technion Israel Institute of Technology\\
\texttt{nailon@cs.technion.ac.il}
}

\iclrfinalcopy 

\begin{document}

\maketitle

\begin{abstract}
Deep learning has proven itself as a successful set of models for learning useful semantic representations of data. These, however, are mostly implicitly learned as part of a classification task.
In this paper we propose the \emph{triplet network} model, which aims to learn useful representations by distance comparisons.
A similar model was defined by Wang et al. (2014), tailor made for learning a ranking for image information retrieval.  Here
we demonstrate using various datasets that our model learns a better representation than that of its immediate competitor, the Siamese network.
We also discuss future possible usage as a framework for unsupervised learning.
\end{abstract}

\section{Introduction}
For the past few years, deep learning models have been used extensively to solve various machine learning tasks. One of the underlying assumptions is that deep, hierarchical models such as convolutional networks
create useful representation of data (\citet{Bengio2009,Hinton2007}), which can then be used to distinguish between available classes.
This quality is in contrast with traditional approaches  requiring engineered features extracted from data and then used in separate learning schemes.
Features extracted by deep networks were also shown to provide useful representation (\citet{Zeiler2013,Sermanet}) which can be, in turn, successfully used for other tasks (\citet{Razavian2014}).  

Despite their importance, these representations and their corresponding induced metrics are often  
treated as side effects of the  classification task, rather than being explicitly sought.
There are also many interesting open question regarding the intermediate representations and their role in disentangling and explaining the data (\citet{Bengio2013}).
Notable exceptions where explicit metric learning is preformed are the \emph{Siamese Network} variants (\citet{bromley1993signature,Chopra2005,hadsell2006dimensionality}), in which a contrastive loss over 
the metric induced by the representation  is used to train the network to distinguish between similar and dissimilar \emph{pairs} of examples.  A contrastive loss favours a small distance between pairs of examples
labeled as similar, and large distances for pairs labeled dissimilar.
However, the representations learned by these models provide sub-par results when used as features for classification, compared with other deep learning models including ours. Siamese networks are also sensitive to calibration in the sense that the notion of similarity vs dissimilarity requires context.  
For example, a person might be deemed similar to another person when a dataset of random objects is provided, but might be deemed dissimilar with respect to the same other person when we wish to distinguish between two individuals in a set of individuals only. 
In our model, such a calibration is not required.
In fact, in our experiments here, we have experienced hands on the difficulty in using Siamese networks.

We follow a similar task to that of \citet{chechik2010large}. For a set of samples $\mathbb{P}$ and a chosen rough similarity measure $r(x,x')$ given through a training oracle
(e.g how close are two images of objects semantically) we wish to learn a similarity function $S(x,x')$ induced by a normed metric.
Unlike  \citet{chechik2010large}'s work, our labels are of the form $r(x,x_1)>r(x,x_2)$ for triplets $x,x_1,x_2$ of objects.
Accordingly, we try to fit a metric embedding and a corresponding similarity function satisfying:
$$ S(x,x_1)>S(x,x_2), \ \ \forall x,x_1,x_2 \in \mathbb{P} \ \ \text{ for which } r(x,x_1)>r(x,x_2).$$
In our experiment, we try to find a metric embedding of a multi-class labeled dataset. We will always take $x_1$ to be of the same class as $x$ and $x_2$ of a different class, although in general more complicated choices could be made.
Accordingly, we will use the notation $x^{+}$ and $x^{-}$ instead of $x_1, x_2$.
We focus on finding an $L_2$ embedding, by learning a function $F(x)$ for which $S(x,x')=\|F(x)-F(x')\|_2$.
Inspired from the recent success of deep learning, we will use a deep network as our embedding function $F(x)$.

We call our approach a \emph{triplet network}.
A similar approach  was proposed in  \cite{WangSLRWPCW2014} for the purpose of learning a ranking function for
image retrieval.  Compared with the single application proposed in  \cite{WangSLRWPCW2014}, we make a comprehensive
study of the  triplet architecture which is, as we shall argue below, interesting in and of itself.
In fact, we shall demonstrate below that the triplet approach  is a strong competitor to the Siamese approach, its
most obvious competitor.

\section{The Triplet network}
A \emph{Triplet network} (inspired by "Siamese network") is comprised of 3 instances of the same feed-forward network (with shared parameters).
When fed with 3 samples, the network
outputs 2 intermediate values - the $L_2$ distances between the embedded representation of two of its inputs from the representation of the third. 
If we will denote the 3 inputs as $x$, $x^{+}$ and $x^{-}$, and the
embedded representation of the network as $Net(x)$, the one before last layer  will be the vector:

\begin{equation*}
    TripletNet(x,x^{-},x^{+})= \begin{bmatrix}
                       \|Net(x)-Net(x^{-})\|_2 \\[0.3em]
                      \|Net(x)-Net(x^{+})\|_2 \\

                      \end{bmatrix} \in \mathbb{R}_{+}^2\ .
\end{equation*}
In words, this encodes the pair of distances between each of $x^+$ and $x^-$ against the \emph{reference} $x$.
\begin{figure}[h]
\begin{center}
\includegraphics[width=0.5\linewidth]{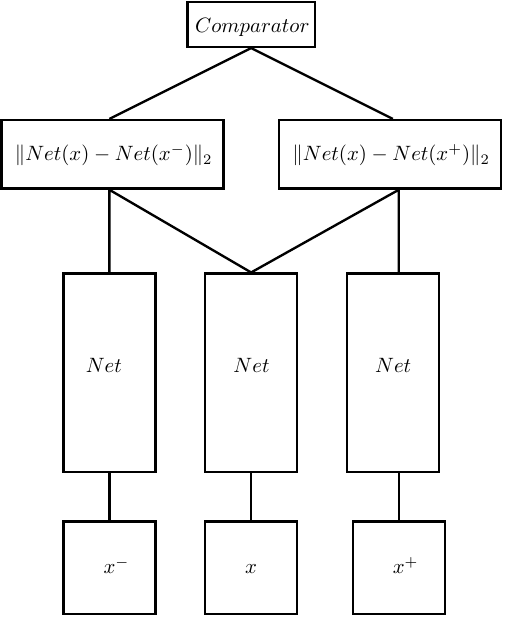}
\end{center}
   \caption{Triplet network structure }\label{tripletnet_scheme}
\end{figure}

\subsection{Training}
Training is preformed by feeding the network with samples where, as explained above, $x$ and $x^{+}$ are of the same class, and $x^{-}$ is of different class.
The network architecture allows the task to be expressed as a 2-class classification problem, where the objective is to correctly classify  which of $x^+$ and $x^-$ is of the same class as $x$.
We stress that in a more general setting, where the objective might be to learn a metric embedding, 
the label determines which example is  \emph{closer} to $x$.
Here we simply interpret ``closeness'' as ``sharing the same label''.
In order to output a comparison operator from the model,
a SoftMax function is applied on both outputs - effectively creating a ratio measure.
Similarly to traditional convolutional-networks, training is done by simple SGD on a negative-log-likelihood loss with regard to the 2-class problem.
We later examined that better results are achieved when the loss function is replaced by a simple MSE on the soft-max result, compared to the $(0,1)$ vector, so that the loss is 
$$Loss(d_{+},d_{-}) = \|(d_{+} , d_{-}-1)\|_2^2 = const\cdot d_+^2$$

where
$$ d_{+} = \frac{e^{\|Net(x)-Net(x^{+})\|_2}}{e^{\|Net(x)-Net(x^{+})\|_2}+e^{\|Net(x)-Net(x^{-})\|_2}}$$
and
$$ d_{-} = \frac{e^{\|Net(x)-Net(x^{-})\|_2}}{e^{\|Net(x)-Net(x^{+})\|_2}+e^{\|Net(x)-Net(x^{-})\|_2}}\ .$$
We note that $Loss(d_{+},d_{-})  \to 0$ iff $\frac{\|Net(x)-Net(x^{+})\|}{\|Net(x)-Net(x^{-})\|} \to 0$, which is the required objective. 
By using the same shared parameters network, we allow the back-propagation algorithm to update the model with regard to all three samples simultaneously.

\section{Tests and results}
 The Triplet network was implemented and trained using the Torch7 environment (\citet{collobert2011torch7}).
\subsection{Datasets}

We experimented with  4 datasets.  The first is \emph{Cifar10}  (\citet{krizhevsky2009learning}), consisting of 60000 32x32 color images of 10 classes (of which 50000 are used for training only, and 10000 for test only). 
The second dataset is the original \emph{MNIST} (\citet{lecun1998gradient}) consisting of 60000 28x28 gray-scale images of handwritten digits 0-9, and a corresponding set of 10000 test images.
The third is the  \emph{Street-View-House-Numbers (SVHN)} of \citet{netzer2011reading} consisting of ~600000 32x32 color images of house-number digits 0-9.
The fourth dataset   is \emph{STL10} of \citet{coates2011analysis}, similar to Cifar10 and consisting of 10 object classes, only with 5000 training images (instead of 50000 in Cifar) and a bigger 96x96 image size.


It is important to note that no data augmentation or whitening was applied, and the only preprocessing was a global normalization to zero mean and unit variance.
Each training instance (for all four datasets) was a uniformly sampled set of 3 images, 2 of which are of the same class ($x$ and $x^{+}$), and the third ($x^{-}$)  of a different class. Each training epoch consisted of 640000 such instances (randomly chosen each epoch), and a fixed set of 64000 instances used for test. We emphasize that each test instance involves 3 images from the set of test images which was excluded from training.
\subsection{The Embedding Net}
For Cifar10 and SVHN we used a convolutional network, consisting of 3 convolutional and 2x2 max-pooling layers, followed by a fourth convolutional layer. A \emph{ReLU} non-linearity is applied between two consecutive layers.
Network configuration (ordered from input to output) consists of filter sizes \{5,3,3,2\}, and feature map dimensions \{3,64,128,256,128\} where a 128 vector is the final embedded representation of the network. Usually in convolutional networks, a subsequent fully-connected
layer is used for classification. In our net this layer is removed, as we are interested in a feature embedding only.

 The network for STL10 is identical, only with stride=3 for the first layer, to allow the bigger input size. The network used for MNIST was a smaller version consisting of smaller feature map sizes \{1,32,64,128\}.

\subsection{Results}
Training on all datasets was done by SGD, with initial learning-rate of 0.5 and a learning rate decay regime. We used a momentum value of 0.9. We also used the dropout regularization technique with $p=0.5$ to avoid over-fitting.
After training on each dataset for 10-30 epochs, the network reached a fixed error over the triplet comparisons.
  We then used the embedding network to extract features from the full dataset, and trained a simple 1-layer network model on the full 10-class classification task (using only training set representations).  The test set was then measured for accuracy.  
  These results (Figure~\ref{results}) are comparable to state-of-the-art results with deep learning models, without using any artificial data augmentation (\citet{zeiler2013stochastic,goodfellow2013maxout,LinCY13}). 
  Noteworthy is the STL10 dataset, in which the TripletNet achieved the best known result for non-augmented data.
  We conjecture that data augmentation techniques (such as translations, mirroring and noising) may provide similar benefits to those described in previous works.

    We also note that similar results are achieved when the embedded representations are classified using a linear SVM model or KNN classification with up to 0.5\% deviance from the results in Figure~\ref{results}.
Another side-affect noticed, is that the representation seems to be sparse - about 25\% non-zero values. This is very helpful when used later as features for classification both computationally and with respect to accuracy, as each class is characterised by only a few non zero elements.

\begin{figure}[h]

    \begin{tabular}{ | c | c | c | p{8cm} |}
    \hline
    Dataset & TripletNet   & SiameseNet  & Best known result (with no data augmentation)       \\ \hline
    Mnist & 99.54$\pm 0.08\%$ & 97.9$\pm 0.1\%$& 99.61\% \citet{mairal2014convolutional,lee2014deeply}  \\ \hline
    Cifar10 & 87.1\% & - & 90.22\% \citet{lee2014deeply}  \\ \hline
    SVHN &  95.37\% & -& 98.18\% \citet{lee2014deeply} \\ \hline
    STL10& 70.67\%& - & 67.9\% \citet{lin2014stable}\\ \hline
    \end{tabular}
     \caption{Classification accuracy (no data augmentation)}\label{results}
     \end{figure}
\subsection{2d visualization of features}
\begin{figure}[h]
\begin{center}
\includegraphics[width=1\linewidth]{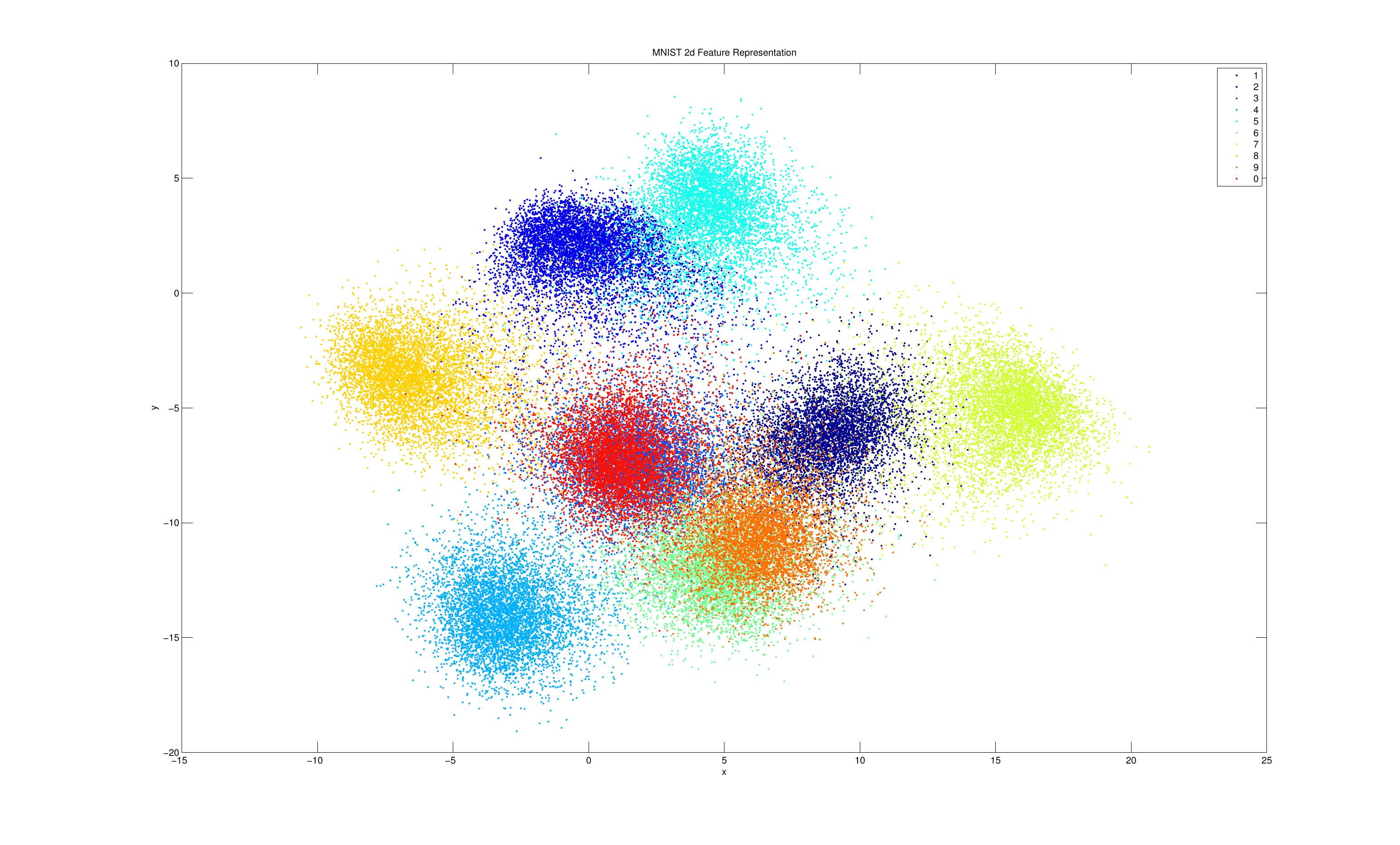}
\end{center}
   \caption{MNIST - Euclidean representation of embedded test data, projected onto top two singular vectors}\label{TripletRepMNIST}
\end{figure}

\begin{figure}[h]
\begin{center}
\includegraphics[width=1\linewidth]{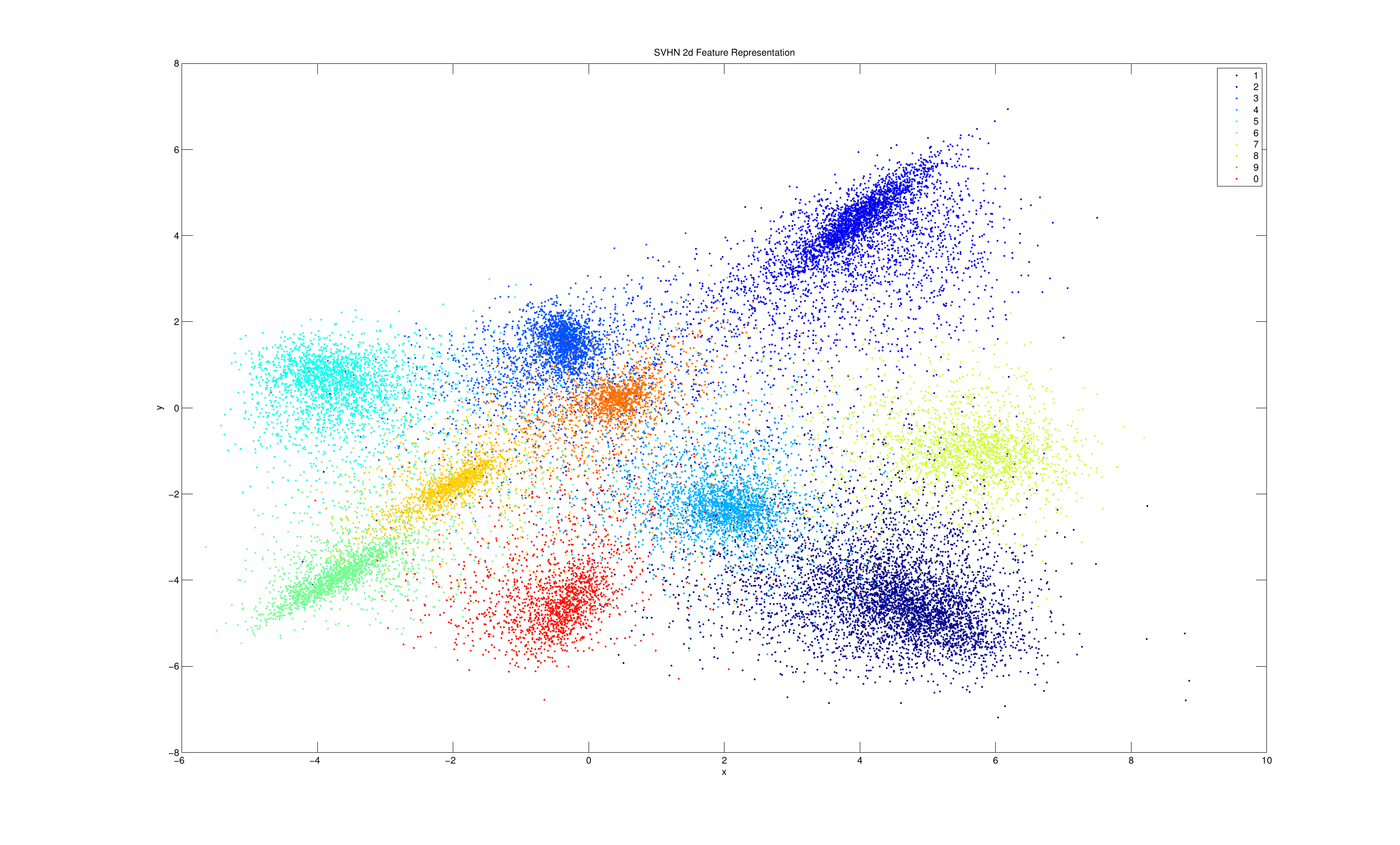}
\end{center}
   \caption{SVHN - Euclidean representation of embedded test data, projected onto top two singular vectors}\label{TripletRepCIFAR10}\label{TripletRepSVHN}
\end{figure}

In order to examine our main premise, which is that the network embeds the images into a representation with meaningful properties, we
use PCA to project the embedding into 2d euclidean space which can be easily visualized (figures \ref{TripletRepCIFAR10} \ref{TripletRepMNIST} \ref{TripletRepSVHN}).
We can see a significant clustering by semantic meaning, confirming that the network is useful in embedding
images into the euclidean space according to their content. 
 Similarity between objects can be easily found by  measuring the distance between their embedding and, as shown in the results, can reach high
classification accuracy using a simple subsequent linear classifier. 

\subsection{Comparison with performance of the Siamese network}
The Siamese network is the most obvious competitor for our approach.  Our implementation of the Siamese network consisted of the same embedding network, but with the use of a contrastive loss between a pair of samples, instead of three (as explained in  \cite{Chopra2005}).  The generated features were then used for classification using a similar linear model as was used for the TripletNet method. We measured lower accuracy on the MNIST dataset compared to results gained using the TripletNet representations \ref{results}. 

We have tried a similar comparison for the other three datasets, but unfortunately could not obtain any meaningful result using a Siamese network.
We conjecture that this might be related to the problem of context described above, and leave the resolution of this 
conjecture to future work.

\begin{figure}[h]
\begin{center}
\includegraphics[width=1\linewidth]{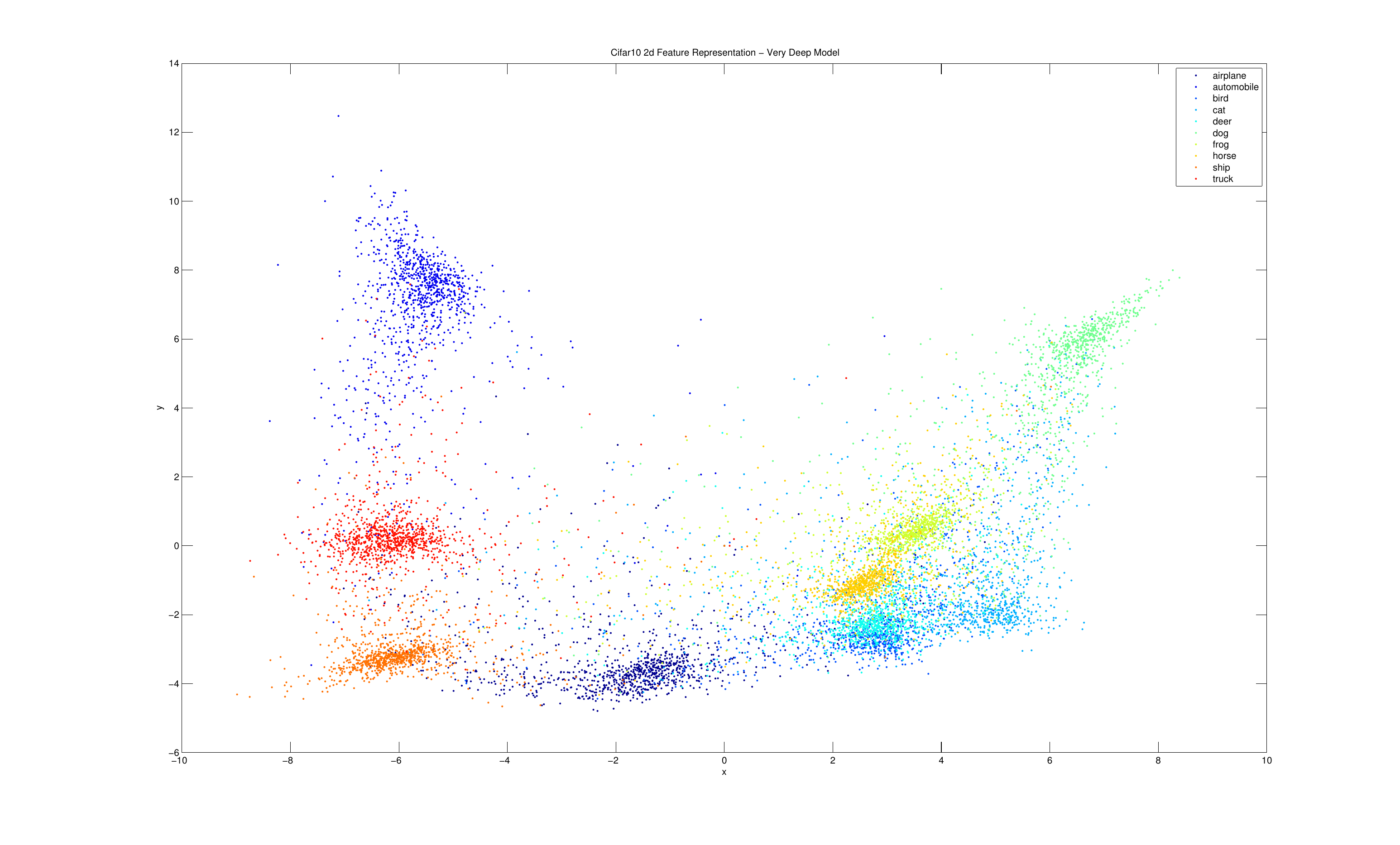}
\end{center}
   \caption{CIFAR10 - Euclidean representation of embedded test data, projected onto top two singular vectors}\label{TripletRepCIFAR10}
\end{figure}
\section{Future work}
As the Triplet net model allows learning by comparisons of samples instead of direct data labels, usage as an unsupervised learning model is possible.
Future investigations can be performed in several scenarios:
\begin{itemize}
 \item {\bf Using spatial information. }Objects and image patches that are spatially near are also expected to be similar from a semantic perspective. Therefore, we could use geometric distance between patches of the same image as a rough similarity oracle $r(x,x')$, in an unsupervised setting.  
\item {\bf Using temporal information.} The same is applicable to time domain, where two consecutive video frames are expected to describe the same object, while a frame taken 10 minutes later is less likely to do so.
Our Triplet net may provide a better embedding and improve on past attempts in solving classification
tasks in an unsupervised environment, such as that of (\citet{mobahi2009deep}).
\end{itemize}
It is also well known that humans tend to be better at accurately providing comparative labels. Our framework can be used in a crowd sourcing learning environment. This can be compared with \citet{shamir}, who used a different approach.
Furthermore, it may be easier to collect data trainable on a Triplet network, as comparisons over similarity measures are much easier to attain (pictures taken at the same location, shared annotations, etc).

\section{Conclusions}
In this work we introduced the \emph{Triplet network} model, a tool that uses a deep network to learn useful representation explicitly.
The results shown on various datasets provide evidence that the representations that were learned are useful to classification in a way that is comparable with a network that was trained explicitly to classify samples. We believe that enhancement to the embedding network
such as Network-in-Network model (\citet{LinCY13}), Inception models (\citet{inception}) and others can benefit the Triplet net similarly to the way they benefited other classification tasks.
Considering the fact that this method requires to know only that two out of three images are sampled from the same class, rather than knowing what that class is, we think this should be inquired further, and may provide us insights
to the way deep networks learn in general.
We have also shown how this model learns using only comparative measures instead of labels, which we can use in the future to leverage new data sources for which clear out labels are not known or do not make sense (e.g hierarchical labels).
\subsubsection*{Acknowledgements}

We gratefully acknowledge the support of NVIDIA Corporation with the donation of the Titan-Z GPU used for this research.

\bibliography{iclr2015}
\bibliographystyle{iclr2015}

\end{document}